\documentclass[conference]{IEEEtran}
\IEEEoverridecommandlockouts
\usepackage{cite}
\usepackage{amsmath,amssymb,amsfonts}
\usepackage{algorithmic}
\usepackage{graphicx}
\usepackage{textcomp}
\usepackage{xcolor}
\usepackage{svg}
\usepackage[utf8]{inputenc}
\usepackage{stfloats} 
\def\BibTeX{{\rm B\kern-.05em{\sc i\kern-.025em b}\kern-.08em
    T\kern-.1667em\lower.7ex\hbox{E}\kern-.125emX}}
\begin{document}

\title{
Guided Visual Attention Model Based on Interactions Between Top-down and Bottom-up Prediction for Robot Pose Prediction
}

\author{
    \IEEEauthorblockN{Hyogo Hiruma}
    \IEEEauthorblockA{
        \textit{Department of Intermedia Art and Science} \\
        \textit{Fundamental Science and Engineering} \\
        \textit{Waseda University}\\
        Tokyo, Japan \\
        hiruma@idr.ias.sci.waseda.ac.jp
    } 
    \and
    
    \IEEEauthorblockN{Hiroki Mori}
    \IEEEauthorblockA{
        \textit{Future Robotics Organization} \\
        \textit{Waseda University}\\
        Tokyo, Japan \\
        mori@idr.ias.sci.waseda.ac.jp
    }
    \and
    
    \IEEEauthorblockN{Hiroshi Ito}
    \IEEEauthorblockA{
        \textit{Controls and Robotics} \\
        \textit{Center for Technology Innovation} \\
        \textit{Research \& Development Group} \\
        \textit{Hitachi, Ltd.}\\
        Ibaraki, Japan \\
        hiroshi.ito.ws@hitachi.com
    }
    \and
    
    \IEEEauthorblockN{Tetsuya Ogata}
    \IEEEauthorblockA{
        \textit{Department of Intermedia Art and Science} \\
        \textit{Fundamental Science and Engineering} \\
        \textit{Waseda University,}\\
        \textit{National Institute of Advanced}\\
        \textit{Industrial Science and Technology (AIST)}\\
        Tokyo, Japan \\
        ogata@waseda.jp
    }
    \and
}

\maketitle

\begin{abstract}
Deep robot vision models are widely used for recognizing objects
from camera images, but shows poor performance when detecting
objects at untrained positions. Although such problem can be
alleviated by training with large datasets, the dataset collection cost
cannot be ignored. Existing visual attention models tackled the problem
by employing a data efficient structure which learns to extract task
relevant image areas. However, since the models cannot modify attention
targets after training, it is difficult to apply to dynamically changing
tasks. This paper proposed a novel Key-Query-Value formulated visual
attention model. This model is capable of switching attention targets by
externally modifying the Query representations, namely top-down attention.
The proposed model is experimented on a simulator and a real-world
environment. The model was compared to existing end-to-end robot vision
models in the simulator experiments, showing higher performance and data
efficiency. In the real-world robot experiments, the model showed high precision
along with its scalability and extendibility.
\end{abstract}

\begin{IEEEkeywords}
neural networks, robotics, visual attention
\end{IEEEkeywords}

\section{Introduction}

Robot task learning incorporating end-to-end learned vision models
enables flexible recognition of complex environments; but has long
suffered by the requirement for large training datasets. Inspired by
the dominant attention models in natural language processing
\cite{nlpattention}, recent studies apply the unique structure for
various purposes, including image recognition  \cite{ViT, CBAM, SelfAttention}
and robot vision processing \cite{DSAE, TransporterNet, TransporterNet2}. 
Attention models for robot vision are designed to reduce the required dataset
size by employing a structure that extracts target data from task relevant
image areas. However, most of such structures have limitations, in which 
the attention targets are intended to be fixed throughout the task.
As real-world robot applications often handle multiple type of objects, the
ability to direct attention to different targets on demand is indispensable.

Originally, attention mechanisms are human cognitive activity to selectively
extract necessary information from innumerable number of stimuli. Generally,
it is considered as an integrated system composed of ``bottom-up attention''
and ``top-down attention'' \cite{HumanAttention}. The bottom-up attention 
automatically collects stimuli of conspicuous areas of the visual receptive
field. By contrast, the top-down attention biases/filters out from the collected
stimuli and selectively extracts a target.

In machine learning, most existing models only consist of visual bottom-up
attention. Although some previous works incorporated controllable visual
attention models \cite{TransporterNet2, ichiwara}, they possess structural
weaknesses that limit application areas and interpretability. The lack of
interpretability is considered to be an critical issue for future robot
applications, such as human-robot collaboration \cite{HRC1, HRC3}.

This work proposed a novel visual attention model, which is capable
of controlling attention targets in a top-down manner. The model was
trained and evaluated on a simulator and real-world environment robot
tasks. The tasks included picking top-down specified objects, which
the model succeeded by inputting extra conditions that explicitly
specify the attention targets. In addition, the simulator results showed the
model's characteristics of high data efficiency, interpretability, and
robustness against appearance changes.

\section{Related Works}

\subsection{Visual processing for robots}
Recognizing object's location and orientation from visual input is an
inevitable task in robot applications. This especially applies for
environment interaction or object manipulation purposes. Studies have
tackled this task by acquiring object-centric representations through various
methods, such as object detection \cite{FasterRCNN, YOLO}, key point
prediction \cite{KeyPoint2, KeyPoint3}, and 3D pose estimation
\cite{3dPose1, 3dPose2}. However, all the methods require a specially
labeled dataset and suffer from its collection cost.

Other studies eliminated the need of specific labels by employing
end-to-end robot task learning \cite{itosan, kasesan}, using
vanilla Fully Convolutional Networks (FCN). End-to-end learning 
enables implicit acquisition of object-centric representations,
by learning to directly map input images and predictions \cite{itosan}.
Despite the elimination of labels, the copious amount of data required
still enlarges the dataset collection cost. One major cause is the
computational nature of FCNs \cite{coordconv}, where the extracted
image features cannot carry the information of where it originated
within the input image. Although it is possible to acquire some positional
information through training, it requires image datasets with objects
placed at various positions. This increases the dataset size, hence
suffer from the collection cost.

\begin{figure*}[t]
 \begin{center}
  \includegraphics[width=0.9\linewidth]{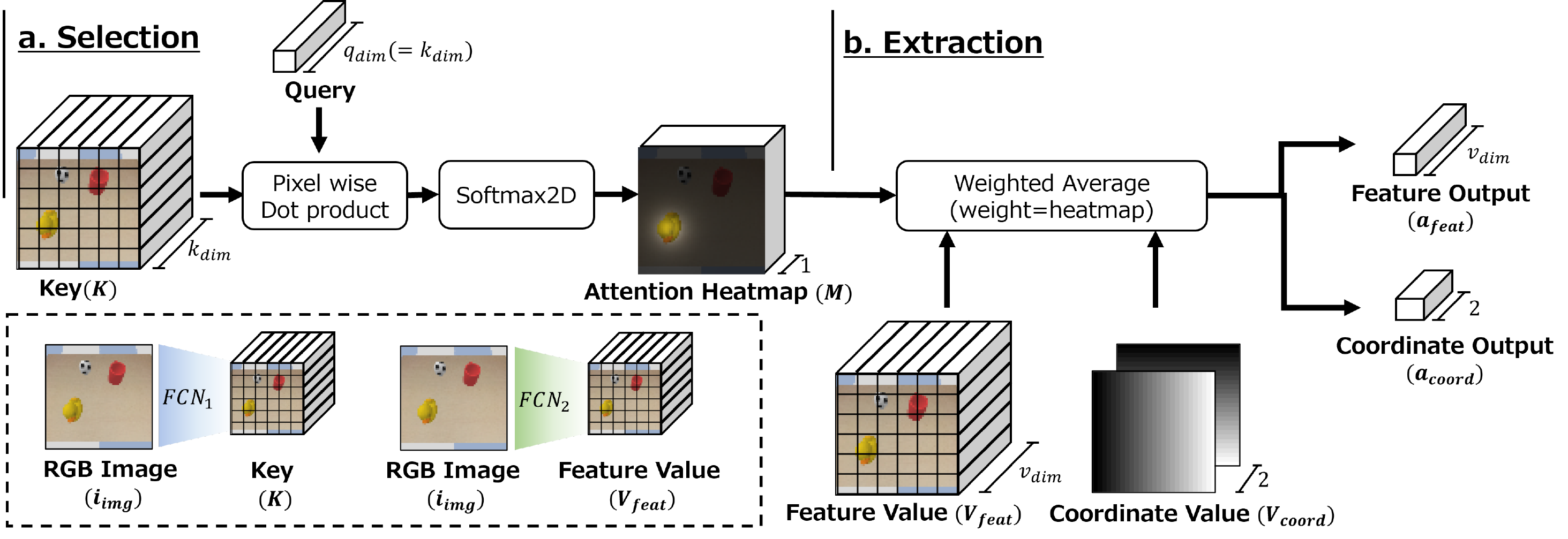}
  \caption{
      Abstract of visual attention model. (a) Selection phase: Generates attention
      heatmap. Scans through the Key and search for the area that matches the Query vector.
      The heatmap is computed by the scaled dot product of the Query vector and the
      feature vector at each pixel of the Key. (b) Extraction phase: Selectively
      extracts feature vectors and coordinates on attended positions. $FCN_{1}$ and
      $FCN_{2}$ do not share the weights.
  }
  \label{fig:model_abstract}
 \end{center}
 \vspace{-3mm}
\end{figure*}

\subsection{Visual attention for robots}

Recent robot vision models \cite{DSAE, TransporterNet, SlotAttention}
utilize attention structures to improve data efficiency. Through end-to-end
learning, attention models learn to predict heatmaps which is composed of
attention weights of each image pixels. The predicted weights tend to
illuminate task relevant targets, such as target objects. In robot tasks,
attention models explicitly extracts the pixel coordinates of where the
objects are located. This eliminates the need of empirically acquiring spatial
representations, and contributes in dataset size reduction.

However, the mentioned visual attention models only employ bottom-up attention,
which cannot modify the attention targets after fixing the model weights.
Although some studies \cite{TransporterNet2, ichiwara} extended the model to
incorporate top-down attentions, they have limitations. Spatial Attention
Point Networks \cite{ichiwara} enabled top-down control by implicitly filtering
out from a redundant number of attention points. Although the model can control
the attention by switching input conditions, the implicit filtering
reduced the model's interpretability. By contrast, the proposed model predicts
on a minumum number of attention points to retain interpretability, where the
targets of each attention points can be modified by specifying with external inputs.
\section{Model Architecture}

\subsection{Spatial attention mechanism}
The proposed attention mechanism model acquires attention through end-to-end
learning, without any explicit supervisions of where or what to attend to. As
shown in Fig. \ref{fig:model_abstract}, the model is constructed using the
Key-Query-Value structure \cite{nlpattention}, and comprises of two phases:
a selection phase (Fig. \ref{fig:model_abstract} (a)) and an extraction phase
(Fig. \ref{fig:model_abstract} (b)). This model can easily be expanded to
multi-head attention by learning multiple weights in parallel. The following
sections will describe the architecture of the selection and extraction phases.

\subsection{Selection phase}
The selection phase scans and predicts the attention weights of each image pixels.
A Key representation $K$ is an image feature of input image $i_{img}$
extracted by a FCN ($FCN_{1}$). $FCN_{1}$ is a three layered convolutional
network with positional embedding concatenated and applied after each layer as
in \cite{coordconv}. The model computes pixel-wise dot product of $K$ and a Query
vector $Q$, and applies 2D softmax with temperature. $Q$ represents an image feature
of the attention target. Such a representation is learned simultaneously by end-to-end
learning; this will be discussed in the later sections. The output is the attention
heatmap $M$, with large weights at pixels where $K$ had image features that are
similar to the $Q$ vector.

\begin{equation}
    M = Softmax_{2D}
        \left(
            \frac{ K \odot Q }{ \sqrt{HW} }
        \right)
\end{equation}
$H$ and $W$ represents the height and width of image feature $K$. $\odot$ stands
for pixel-wise scaled dot product \cite{nlpattention}. During above operation, the
roles of $K$ and $Q$ are to mutually interact to construct task relevant
representations. In this sense, $K$ and $Q$ act as the bottom-up and top-down
attention, respectively.

\subsection{Extraction phase}
The extraction phase extracts the image feature and coordinate of attended
locations, in reference to $M$. First, the model generates two Value
representations that spatially correspond to $M$: Feature Value $V_{feat}$
and Coordinate Value $V_{coord}$. $V_{feat}$ is an image feature of the input
image, converted with $FCN_{2}$ which does not
share weights with $FCN_{1}$. $V_{coord}$ is an absolute positional embedding
which lists the coordinate values of each pixel in image format.
For simplicity, this model employs a Cartesian coordinate embedding which
lists x and y coordinates in two different channels.

Given these value representations, the model calculates a weighted average
through spatial dimension, or expectation operation as referred to in \cite{DSAE}.
The outputs are treated as the selective extraction of the attention model: target
feature vector $a_{feat}$ and target coordinate $a_{coord}$. The extracted data will
contain values that are close to those that originated at the attended pixels, 
since $M \in [0.0, 1.0]$. $(u, v)$ represents the position of each pixels.
\begin{equation}
    a_{feat} = 
        \sum_{u=0}^{W}
        \sum_{v=0}^{H}
        M_{(u, v)} V_{feat (u, v)}
\end{equation}
\begin{equation}
    a_{coord} =
        \sum_{u=0}^{W}
        \sum_{v=0}^{H}
        M_{(u, v)} V_{coord (u, v)}
\end{equation}

\begin{figure}
    \begin{center}
        \includegraphics[width=\linewidth]{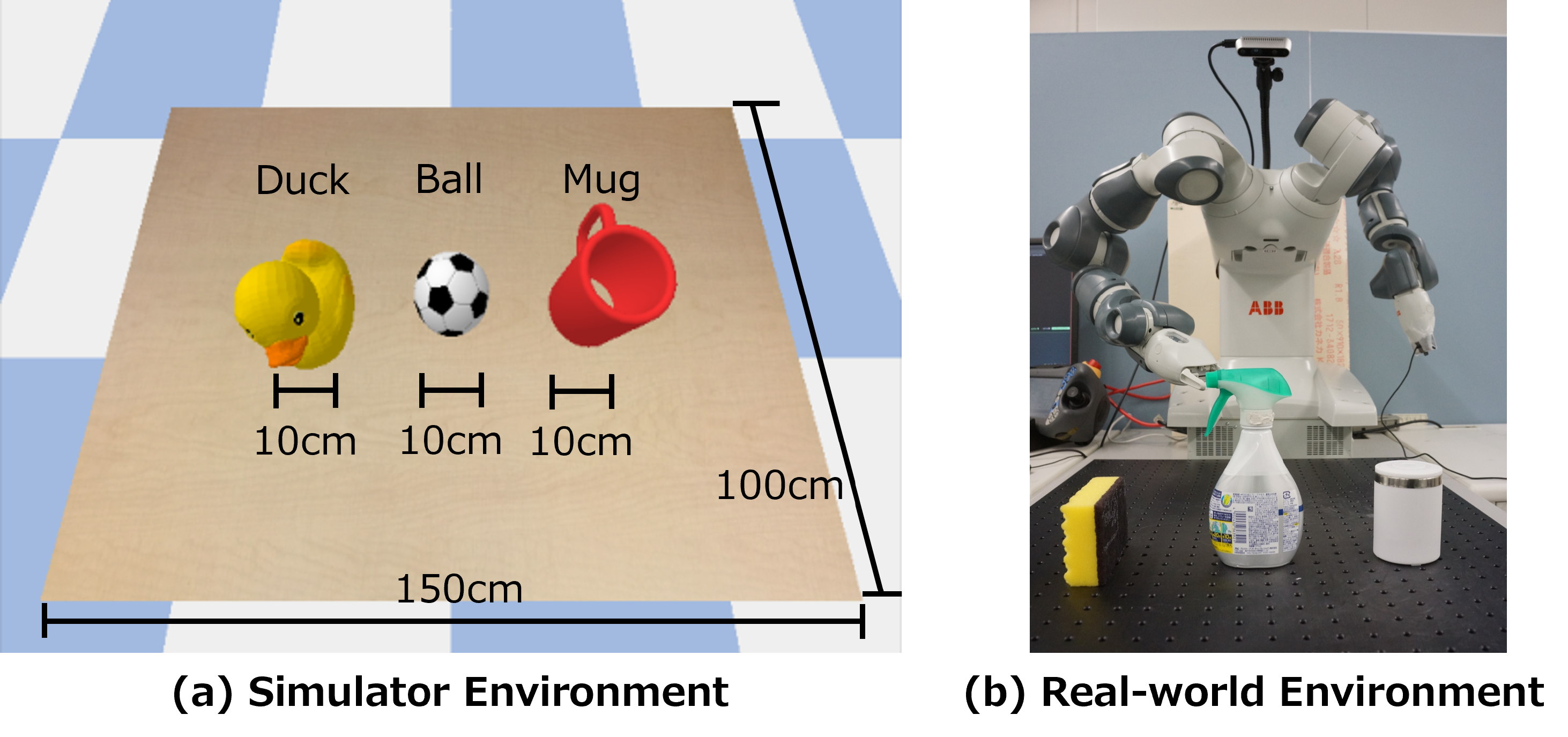}
        \caption{Environments: (a) A simulator environment running on PyBullet,
        including three types of figures. (b) Real-world environment contains a
        7DOF dual-arm robot with three types of objects.
        }
        \label{fig:envs}
    \end{center}
    
    \begin{center}
        \includegraphics[width=\linewidth]{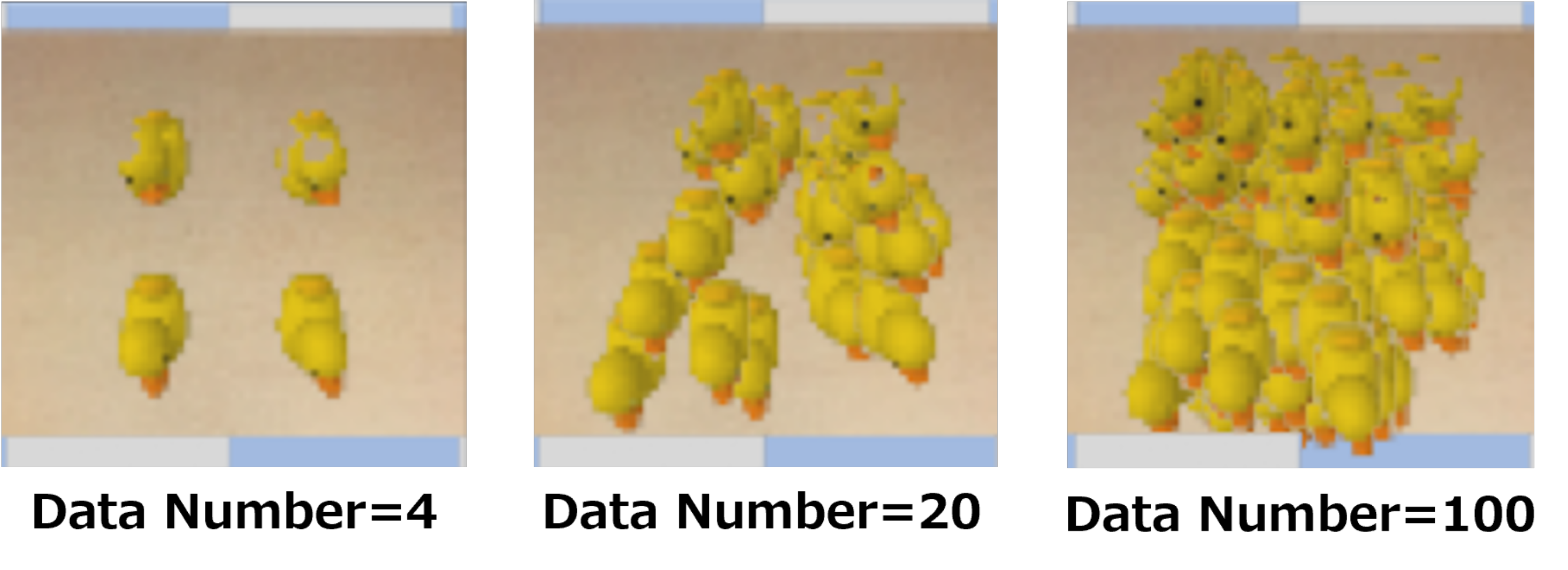}
        \caption{
          Relation of positional density and dataset size. Each
          image displays the total area and density covered by the object.
          The dataset is based on a task (a). Above images are rendered for
          visualization.
        }
        \label{fig:sim_dataset}
    \end{center}
     \vspace{-2mm}
\end{figure}

\subsection{Query vector generation and top-down attention}
One strength of the proposed model is that it is structured to store object-centric
representations in a distinct variable $Q$, which existing models cannot.
This allows the model to modify the attention targets by switching $Q$s with
corresponding representations. Such $Q$s can be created in various ways, depending
on the task. This subsection will describe two methods.

\subsubsection{Base Query}
A simple implementation is to use learnable variables as $Q$s. The variables
are initialized with random values and are trained to store target
object representations; we term such a variable as Base Queries $q_n^B$.
Multiple Base Queries can be learned simultaneously and be manually switched
on inference. This way, one can modify the attention targets in a top-down manner.

\begin{table}
    \centering
    \caption{Experiment task configurations}
    \scalebox{1}{
        \begin{tabular}{l|c|c|c}
            \hline
                                  & Number of & Number of & Top-down      \\
            Task name             & Presented & Target    & Specification \\
                                  & Objects   & Objects   & \\ \hline
            (a) Single Object     & 1         & 1         & - \\
            (b) Multiple Object   & 3         & 3         & - \\
            (c) Selected Type     & 3         & 1         & Object type \\
            (d) Selected Position & 2         & 1         & Object Position \\ \hline
        \end{tabular}
    }
    \label{tab:exp_config}
    \normalsize
     \vspace{-2mm}
\end{table}

\subsubsection{Conditioned Query}
Conditioned Queries are Queries created based on additional input conditions
$c_{query}$. The conditions include one-hot vectors, images, natural language
embeddings, which are embedded to vector representations using multi-layer
perceptrons (MLP). This can be combined with Base Queries, for learning different
targets with similar appearances (e.g. attending to one of identical objects
based on their positions). The Base Queries will store the major visual
characteristics and the conditions will support to distinguish the minor
differences.
\section{Experiments}
We experimented to verify the features of the proposed model on simulator and
real-world environments. The simulator experiment trained the model on an object
position prediction task.  Preciseness and sample efficiency were evaluated using
datasets with different sizes (Fig. \ref{fig:sim_dataset}), along with
comparisons against existing models. Whereas the real-world experiment focused
on the proposed model and verified if it was applicable to noisy robot task settings.
We trained it to predict object picking poses, which requires both image feature and
coordinate data of attention points to determine how and where to perform the picking
motion.

\subsection{Environment configurations}
For the simulator experiment, we employed a physics simulator called
PyBullet \cite{PyBullet} for environment simulations. As shown in
Fig. \ref{fig:envs}(a), it was composed of a chequered floor and wooden textured
table along with three types of objects: a duck figure, soccer ball, and mug. Each
object was resized to a width of approximately 10 cm. The virtual camera
was set to focus on the center of the table from a distance of 50 cm.

For the real-world experiment, we used IRB 14000 YuMi from ABB. This is a human
collaborative dual-arm robot with seven degrees of freedom each and a gripper.
As shown in Fig. \ref{fig:envs}(b), three objects were used for the tasks:
a sponge, spray, and mug. A Realsense D435 RGBD camera was used for collecting RGB
images, which was installed on top of the robot body. 

\begin{table*}
    \centering
    \caption{\textbf{Simulator experiment:} Average error distance in centimeters. Bolded below 5cm}
    \scalebox{1}{
        \begin{tabular}{l|c c c|c c c|c c c|c c c} \hline
                              & \multicolumn{6}{c|}{Baseline Task} & \multicolumn{6}{c}{Top-down Task}\\
            \hline
            Prediction Target & \multicolumn{3}{c|}{Single Object} & \multicolumn{3}{c|}{Multiple Objects} & \multicolumn{3}{c|}{Selected Type} & \multicolumn{3}{c}{Selected Position}\\
            \hline
            Dataset size      & 4 & 20 & 100 & 4 & 20 & 100 & 4 & 20 & 100 & 4 & 20 & 100 \\
            \hline
            Ours (q=min)                          & \textbf{4.14} & \textbf{2.88} & \textbf{3.83}
                                                  & \textbf{4.80} & \textbf{3.21} & \textbf{2.80}
                                                  & 27.16         & \textbf{4.27} & \textbf{4.62}
                                                  & 32.81         & \textbf{4.85}    & \textbf{3.95} \\
            Key Point (q=16) \cite{KeyPoint3}     & \textbf{2.16} & \textbf{3.69} & \textbf{0.62}
                                                  & 22.41         & \textbf{3.85} & \textbf{2.61}
                                                  & 33.28         & 24.61         & 7.94
                                                  & 31.82         & 15.75         & 9.63 \\
            Key Point (q=min) \cite{KeyPoint3}    & 18.79         & 18.07         & \textbf{0.65}
                                                  & 41.90         & 48.47         & 49.02
                                                  & 38.90         & 23.54         & 5.68
                                                  & 30.10         & 20.51         & 9.17 \\
            Deep Spatial AE (q=16) \cite{DSAE}    & 15.12         & 15.52         & 60.05
                                                  & 22.69         & 29.18         & 30.26
                                                  & 27.12         & 39.30         & 28.55
                                                  & 43.97         & 36.22         & 28.41\\
            Convolutional AE                      & 34.07         & 30.81         & 17.17
                                                  & 65.84         & 44.16         & 46.88
                                                  & 43.84         & 40.21         & 39.98
                                                  & 81.79         & 19.88         & 40.88 \\
            FCN                                   & 14.91         & 6.18          & \textbf{1.60}
                                                  & 33.94         & 31.70         & 30.52
                                                  & 26.74         & 27.62         & 28.53
                                                  & 34.82         & 19.42         & 14.82 \\
            \hline
        \end{tabular}
    }
    \label{tab:exp1_results}
    \normalsize
     \vspace{-3mm}
\end{table*}

\subsection{Task settings}
The general objective on both environments was to learn to recognize an object
location. As shown in Table \ref{tab:exp_config}, we derived four different tasks
to break down and evaluate the ability. Tasks (a) and (b) comprised the baseline tasks
and tasks (c) and (d) comprised the top-down tasks. Baseline tasks evaluates the model's
data efficiency. We trained the model to recognize the positions of all visible objects.
In contrast, top-down tasks evaluated the model's ability to predict positions of only
a top-down specified target. Task (c) specified the object type, and task (d) specified
whether the object is placed on the left or right side of the table.

\subsection{Training configurations}
The proposed model was trained to predict a minimum number of attention points
that match the number of predicting objects (Table \ref{tab:exp_config}).
For query generation, the Base Queries and Conditioned Queries were used for
the baseline tasks and top-down tasks, respectively. MLPs were used to convert
attention coordinate values to object coordinates in simulator environment.
Similarly, attention coordinate and feature values were concatenated and converted
to robot picking poses with MLPs. The simulator and real-world tasks were trained
to minimize the errors between the prediction and ground truth coordinate and robot
pose values, respectively.

\begin{figure*}[htbp]
    \begin{center}
        \includegraphics[width=0.9\linewidth]{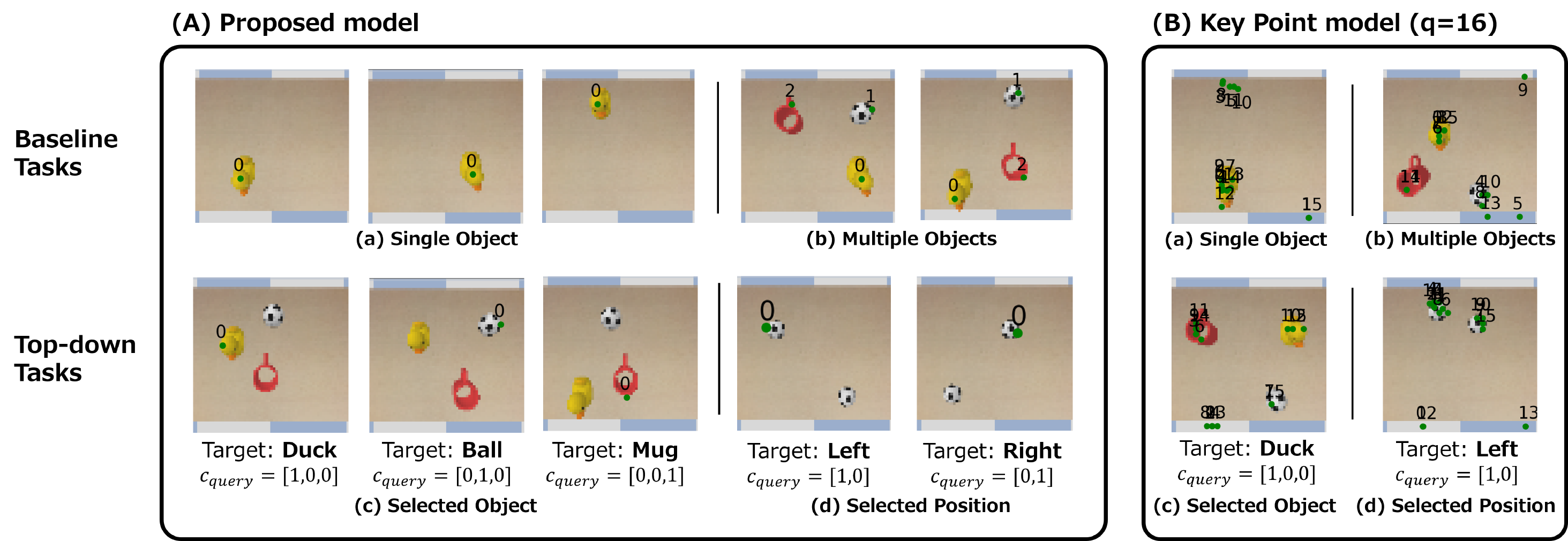}
        \caption{
            (A) Visualization of predicted attention of the proposed model on simulator
            experiments. The green dots represent attention points, along with
            attention index numbers. (a)(b) Results of baseline tasks. (c)(d) Results of
            Top-Down tasks, each with different input conditions. (B) Visualization of
            attentions predicted by Key Point model (q=16) \cite{KeyPoint3}.
        }
        \label{fig:sim_results}
    \end{center}
     \vspace{-3mm}
\end{figure*}

\begin{figure*}
 \begin{center}
  \includegraphics[width=0.9\linewidth]{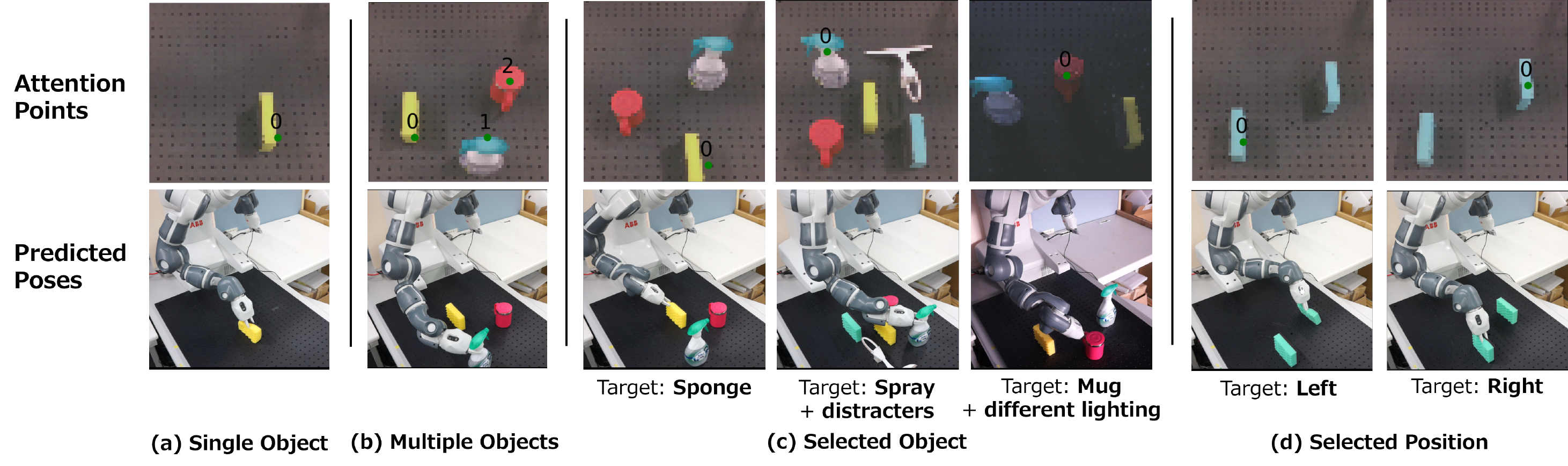}
  \caption{
    Prediction results of real-world experiments. Top-row: visualization
    of attention points. Bottom-row: predicted robot poses for picking
    target objects.
  }
  \label{fig:real_result}
 \end{center}
 \vspace{-3mm}
\end{figure*}

The proposed model was compared with existing visual attention models and
vanilla FCN based models. The visual attention models included ``KeyPoint''
model \cite{KeyPoint3} and ``Deep Spatial AE (Autoencoder)'' \cite{DSAE}.
Other attention models \cite{TransporterNet2, ViT} were excluded because
they required specialized annotation or they are irrelevant to improving data
efficiency. ``Key Point'' model \cite{KeyPoint3} is composed of a vanilla
FCN followed by a spatial softmax layer. Each image feature channel of the
output is referred to as an attention heatmap and the coordinates of the most
weighted pixels are output as the attention points. The model is tested on two
conditions, in which the model was trained with the minimum (q=min) and
redundant (q=16) number of attentions; ``q'' denotes the number of learned
attention points. ``Deep Spatial AE (Autoencoder)'' \cite{DSAE} is an hourglass
encoder-decoder model which employs the ``KeyPoint'' model as an encoder. In
task (c) and (d), the one-hot vectors, that specify the targets, were input
with the coordinates to convert to desired outputs using MLPs.

The vanilla FCN based models included ``Convolutional AE'' and vanilla ``FCN''.
``Convolutional AE'' is an hourglass encoder-decoder model that employs
convolutional layers and MLPs to compress an RGB image into a single
vector. The compressed vector was input to MLPs for succeeding coordinate
prediction.
\section{Experiment Results}

\subsection{Simulator experiment}
Table \ref{tab:exp1_results} shows the experiment results on the simulator
environment, along with the comparisons against existing vision models. The
values are the average error distance between predicted and ground truth object
coordinates. We bolded the values that had errors below 5cm, which is the half
of the object width. Each model were evaluated on predictions with objects at
100 random positions. 

\subsubsection{Proposed model}
Table \ref{tab:exp1_results} shows that the proposed model predicted object
coordinates with a slight error under most conditions, when compared with
other models. It is notable that the model retained high performance even
on minimum training configurations, such as dataset size and the number
of attentions. The predicted attention points on the untrained data are
displayed in Fig. \ref{fig:sim_results} (A). The points accurately located on
the objects, which also suggests that the model adapted to appearance changes
affected by the relative position to the camera. The proposed model succeeded
on both top-down tasks (c) and (d), which indicates the model's ability to
specify targets based on appearances and spatial representations, namely
feature-based attention and position-based attentions. However, when trained on
dataset of size 4, the model failed to structure the queries for all target
objects. The model predicted odd attention points, such as failing to acquire
attention for one of three objects in task (c).

\subsubsection{Comparison models}
By contrast, Table \ref{tab:exp1_results} shows that Key Point models have lower
sample efficiency: task (a) and (b) for KeyPoint (q=min) and task (b) for KeyPoint
(q=16). Although Key Point models scored the highest when trained on the largest
dataset in task (a), Fig. \ref{fig:sim_results} (B) shows that some points attended
to irrelevant locations, which decreases the interpretability. In addition, Fig.
\ref{fig:sim_results} (B) shows multiple points attending to different parts of a 
single object. This is because each attention heatmap only considers a single visual
pattern, thus becomes vulnerable against minor appearance changes. Thus, the redundant
number of attention points were necessary to increase the number of considered visual
patterns.

Convolutional AE and Deep Spatial AE failed to predict on both of the baseline
tasks. The reconstructed images showed images with background but no target
objects. Considering the small space a single object occupied on the environment,
the model is likely to have misinterpreted the object as a noise. This result suggests that
separately training vision models induces a negative impact, especially when
trained on dataset with trivial visual stimulus. The results of FCN models
indicated an increase in accuracy when provided with sufficiently dense datasets,
which was as expected, considering the density of object positions in large
datasets (Fig. \ref{fig:sim_dataset}).

All the existing models showed low precision on both top-down tasks. The Key Point
model (q=16) had the best score among the compared models, which successfully
attended to the presented objects (Figs. \ref{fig:sim_results} (B) (c) and (d)).
However, the model had less precision even when given a sufficient number of
datasets and attention points. This is likely to be caused by the instability of
the predicted attention points.

\subsection{Real-world experiment}

Table \ref{tab:exp2_results} shows the success rate of picking an object, each
evaluated on 30 trials, where the target objects were placed at untrained positions.
In Fig. \ref{fig:real_result}, the top row shows the camera image with predicted
attention points, and the bottom row shows the predicted robot poses. The model
succeeded to direct the attention points to each target objects and to implicitly map
the attention coordinates to real-world 3D coordinates to perform the picking motion.
As shown in Fig. \ref{fig:real_result} (c), the model predicted different picking
poses or approaching directions depending on the target object type. This result
indicates that the model succeeded to distinguish the object targets based on
the jointly extracted image features. Although object types are actually given 
as conditions in this task, the ability to change predictions based on learned
image features is essential to extend this model in the future works. The model
also showed high robustness against noises, such as when presented with similar
irrelevant objects or with darker lighting environments (Fig. \ref{fig:real_result}
(c)). On the few failed trials (Table \ref{tab:exp2_results}), the model did succeed
in predicting appropriate attention points, but the predicted gripper position tended
to shift slightly. This occurred when the object appearance was affected by a strong
distortion of the camera lenses, especially when placed at the edge of the environment.

\begin{table}
    \centering
    \caption{\textbf{Real-world Experiment:} Success rate of the object picking}
    \scalebox{1}{
        \begin{tabular}{l|c|c|c|c} \hline
                  & \multicolumn{2}{c|}{Baseline Task} & \multicolumn{2}{c}{Top-down Task}\\
            \hline
            Task & Single & Multiple & Selected & Selected \\
                 & Object & Objects  & Type     & Position\\
            \hline
            Dataset size      & 20 & 20 & 20 & 20 \\
            \hline
            Success Rate & \textbf{27}/30 & \textbf{28}/30 & \textbf{26}/30 & \textbf{27}/30 \\
            \hline
        \end{tabular}
    }
    \label{tab:exp2_results}
    \normalsize
     \vspace{-3mm}
\end{table}

\section{Conclusion}
This paper presented a novel end-to-end learned top-down visual attention
model. By separating the target representation to an independent Query vector,
the model was able to guide the attention to different targets based on external
input conditions. Experiments on simulation and real-world environments suggested
that the model performed with high precision and interpretability even with minimum
training configurations. In addition, introducing position-based attention and the
feature extraction showed wider application capacity. However the model has limitations,
where the model cannot automatically self-control the attention targets.
Considering real-time robot controlling tasks, interaction targets are expected
to change as the task proceeds. Therefore, we are currently working  on expanding
this model to employ such active top-down attention, in contrast to current
static top-down attention.

\section*{Acknowledgment}
\addcontentsline{toc}{section}{Acknowledgment}
This work was supported by JST Moonshot R\&D Grant Number JPMJMS2031 and
by JSPS KAKENHI Grant Number JP21H05138.

\bibliography{bibliography}
\bibliographystyle{unsrt}

\end{document}